\documentclass[letterpaper, 10 pt, conference]{ieeeconf}
\IEEEoverridecommandlockouts

\let\labelindent\relax

\usepackage[T1]{fontenc}

\usepackage[inline]{enumitem}
\usepackage{amsmath}
\usepackage{amssymb}
\usepackage{multirow}
\usepackage{graphicx}
\usepackage[font=small]{caption}
\usepackage{subcaption}
\usepackage[normalem]{ulem}
\usepackage[dvipsnames]{xcolor}
\usepackage{xcolor}
\usepackage{colortbl}
\usepackage{url}
\usepackage{dsfont}
\usepackage{tikz}\usetikzlibrary{arrows.meta, positioning, fit, backgrounds, calc}

\newcommand{\eg}{\textit{e.g.,}}

\newif\ifdoubleblind
\doubleblindfalse

\usepackage{xcolor}

\newcommand{\blindtext}[2][anonymized]{%
  \ifdoubleblind
    \textcolor{blue}{[#1]}%
  \else
    #2%
  \fi
}

\newif\ifpresistverboseloss
\presistverboselosstrue    

\newif\ifverbosecitations
\verbosecitationsfalse    

\makeatletter
\def\bstctlcite{\@ifnextchar[{\@bstctlcite}{\@bstctlcite[@auxout]}}
\def\@bstctlcite[#1]#2{\@bsphack
  \@for\@citeb:=#2\do{%
    \edef\@citeb{\expandafter\@firstofone\@citeb}%
    \if@filesw\immediate\write\csname #1\endcsname{\string\citation{\@citeb}}\fi}%
  \@esphack}
\makeatother

\title{
\LARGE \bf
PreSIST: Vision-Language-Informed Object Persistence Prediction in Open-World Scenes
}
\newcommand{\modifiedAAA}[1]{#1}

\newcommand{\cutDL}[1]{}
\newcommand{\cutAAA}[1]{}
\newcommand{\lengthCutAAA}[1]{}

\author{Amanda Adkins \and Tarunvidyut Ravisankar \and Joydeep Biswas
\thanks{This work has taken place in the Autonomous Mobile Robotics Laboratory (AMRL) at UT Austin. AMRL research is supported in part by National Science Foundation (CAREER-2046955, OIA-2219236, DGE-2125858, CCF-2319471, GRFP DGE-2137420), ARO (W911NF-23-2-0004), Amazon, and JP Morgan. Any opinions, findings, and conclusions expressed in this material are those of the authors and do not necessarily reflect the views of the sponsors.}
\thanks{The authors are with the Department of Computer Science, The University of Texas at Austin, Austin, TX. Email: {\tt\small \{aaadkins, tarunrav,  joydeepb\}@cs.utexas.edu }}
}

\begin{document}
\maketitle
\IEEEpeerreviewmaketitle

\ifverbosecitations
  \bstctlcite{BSTcontrol:verbose}
\else
  \bstctlcite{BSTcontrol:terse}
\fi


\begin{abstract}
Robots deployed over long periods must reason about environments that change over time. Existing long-term perception systems often address object change reactively, updating their maps only after revisiting a scene and observing that an object has moved.
Instead, robots should reason proactively about how long objects are likely to persist using the context in which they appear. For example, a car at a traffic light and a car in a parking spot share the same semantic class, but their contexts imply different persistence durations.
We propose \textbf{PreSIST} (\textbf{Pre}dictive \textbf{S}cene-conditioned \textbf{I}nstance \textbf{S}urvival over \textbf{T}ime), a method for predicting whether an observed object will remain in its last seen pose at arbitrary future times. PreSIST estimates instance-level persistence priors from object properties and scene context, then integrates these priors with a probabilistic persistence filter as observations become available.
Its key insight is that the reasoning capabilities of vision-language models (VLMs) can relate scene context to likely object use and human activity, enabling persistence prediction before long-term observations are available.
We develop two interchangeable variants: \textbf{PreSIST-Lang}, which estimates persistence priors using a VLM, and \textbf{PreSIST-Vis}, a novel vision-only model trained using PreSIST-Lang pseudo-labels for efficient deployment.
Experiments on a new dataset of in-the-wild object persistence annotations show that PreSIST-Lang and PreSIST-Vis outperform baselines on open-world persistence prediction.
Our dataset and code are available at \blindtext{\url{https://github.com/ut-amrl/PreSIST}}.


\end{abstract}



\section{Introduction}
\label{sec:intro}







Robots operating over long periods must reason about how environments change over time. While many SLAM and perception systems assume a static world~\cite{wang2024survey}, real-world objects are often only temporarily stable: parked cars drive away,\cutAAA{ packages appear and disappear,} pallets are moved, and chairs are rearranged. These objects may be useful when first observed, but later become unreliable for mapping, localization, and planning. 
Existing systems usually address this problem reactively~\cite{sousa2023systematic, Schmid-RSS24-Khronos}, updating only after collecting enough \cutAAA{contradictory }evidence to conclude that an object has moved. Until then, the robot may \cutAAA{continue to }rely on outdated information.
An alternative is to assign dynamics by semantic class, either filtering out assumed movable classes or using class-level heuristics about persistence. However, semantic class alone is insufficient for predicting how long an object will remain stationary. A car at a traffic light is likely to \cutAAA{leave its observed pose}\modifiedAAA{move} within tens of seconds, whereas a car in a parking spot may remain \cutAAA{in the same pose }stationary for hours. Both belong to the
same class, but their persistence depends on scene context, human activity, and the object's
role in the environment.

\begin{figure}[tbp]
    \centering
    \includegraphics[width=0.77\columnwidth]{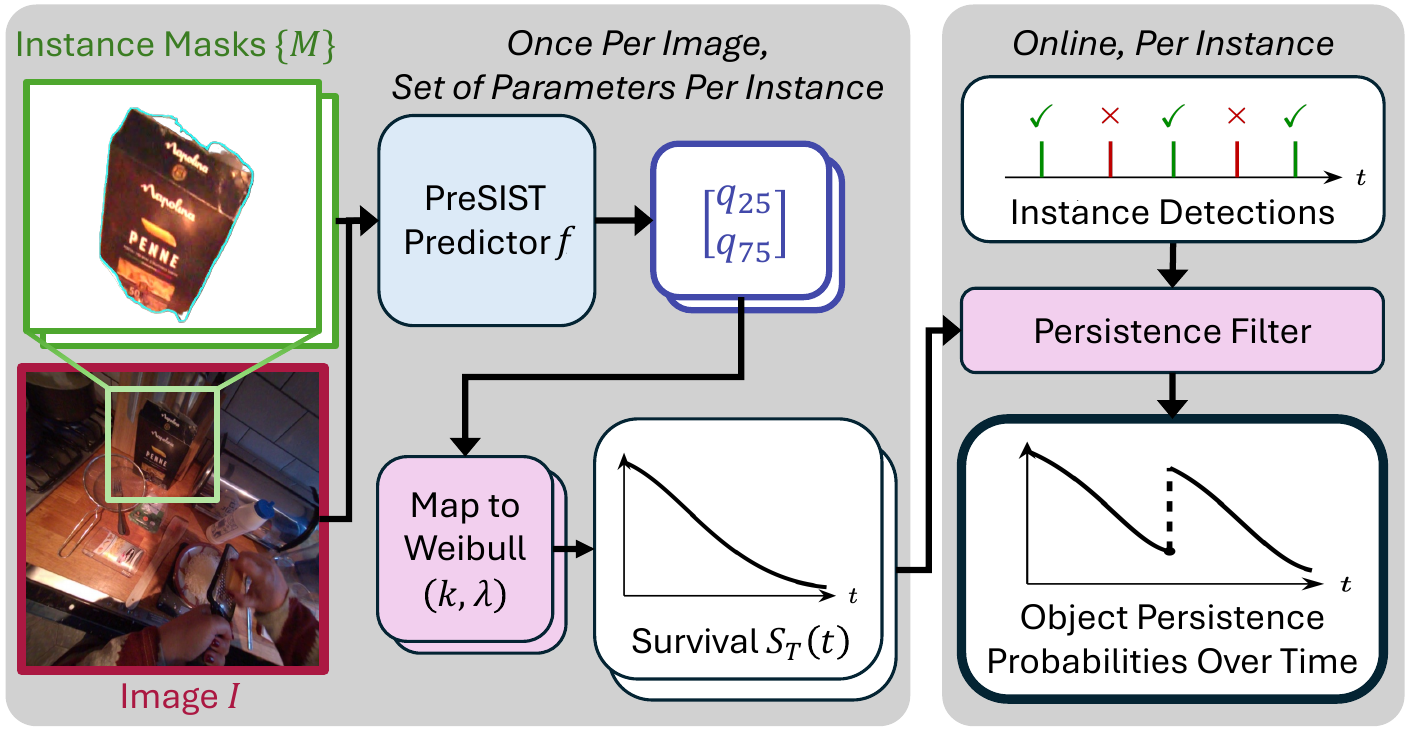}
    \caption{PreSIST Overview. From an  image $I$ and instance masks $\{M\}$,
    the predictor $f$ emits per-object survival-time quantiles $(q_{25}, q_{75})$ defining a Weibull survival prior $S_T(t)$. A persistence filter combines this with online positive and negative detections to estimate the probability that the object is still present over time.}
    \label{fig:presist_overview}
\end{figure}

In this paper, we ask whether a robot can proactively estimate object persistence: how long an observed object will remain in its last seen pose. Specifically, rather than waiting for evidence of change to accumulate, we seek to predict how long the object is likely to remain stationary based on object identity and scene context. To this end, we propose \textbf{PreSIST} (\textbf{Pre}dictive \textbf{S}cene-conditioned \textbf{I}nstance \textbf{S}urvival over \textbf{T}ime), a method for estimating scene-conditioned object persistence. PreSIST uses object properties and surrounding scene context to estimate object-level persistence priors, then integrates these priors with a probabilistic persistence filter~\cite{rosen2016towards}, as depicted in Fig.~\ref{fig:presist_overview}. The persistence filter maintains a belief over whether an observed object will remain in its last seen pose at an arbitrary future time $t$, updating that belief as new observations arrive. The object-level persistence prior initializes the filter's belief, while the filter provides efficient time-indexed prediction and principled incorporation of future evidence. We develop two interchangeable variants of PreSIST. \textbf{PreSIST-Lang} estimates persistence priors using a large vision-language model (VLM), leveraging such models' reasoning abilities related to object use, human activity, and scene context~\cite{chu2024timebench, yuan2025eocbench, liu2024mmbench}. \textbf{PreSIST-Vis} is a vision-only model trained from PreSIST-Lang pseudo-labels, enabling efficient deployment without online large-model inference while retaining the context-sensitive persistence estimates provided by the language-based model.


Prior persistence-filtering and related approaches typically obtain persistence models either by estimating them from long-term in-context observations~\cite{saavedra2025perpetua, krajnik2017fremen, rotsidis2021exmaps, symmank_spatio-temporal_2023}, which can require difficult-to-collect data that may not cover the diversity \cutAAA{of objects, environments, and contexts }encountered in long-term deployments, or by specifying them from semantic categories and heuristics~\cite{rosen2016towards, bogenberger2026glasses}, which can miss contextual cues that affect persistence. In contrast, PreSIST obtains context-conditioned priors without requiring long-term observations of the target environment or category-level specification, while preserving the key benefits of persistence filtering: probabilistic predictions of object persistence, efficient evaluation at arbitrary future times, and responsiveness to new observations.

We summarize our contributions as follows:
\begin{enumerate*}[label=(\roman*)]
\item a method, PreSIST-Lang, for estimating persistence priors from object properties and scene context using a large vision-language model, enabling persistence filters to produce probabilistic object-persistence predictions at arbitrary future times in unseen, open-world settings;
\item a vision-only model, PreSIST-Vis, that approximates these foundation-model-based persistence estimates without requiring online large-model inference;
\item a manually labeled dataset of object presence intervals and masks, built on existing and newly collected image sequences, for assessing context-aware persistence estimation.
\end{enumerate*}
We empirically show that PreSIST-Lang and PreSIST-Vis outperform baselines of comparable efficiency on open-world persistence prediction.

\section{Related Work}
\label{sec:related_work}
%
%

\subsection{Long-Term Mapping and Object Persistence Modeling}
%
Robotic systems deployed in dynamic environments must account for object movement, yet most long-term
perception systems treat such change
reactively, updating their representation of the scene only once observations contradict
it~\cite{sousa2023systematic, Schmid-RSS24-Khronos}.
A complementary body of work instead predicts an object's or feature's future state. Some methods use periodic models of environment dynamics~\cite{krajnik2017fremen};
others forecast object-level changes from symbolic sequences of object
arrangements~\cite{patel2023slatepro}.
A common predictive strategy draws on survival analysis, modeling the probability that a
feature remains as a function that decays with the time since it was last seen: Toris
and Chernova~\cite{toris2017temporal} and ExMaps~\cite{rotsidis2021exmaps} use
exponential decay, and Symmank~\cite{symmank_spatio-temporal_2023} fits a Weibull
survival model via recursive Bayesian estimation. Persistence filters, introduced by Rosen et al.~\cite{rosen2016towards}, embed a survival
model in a recursive Bayesian estimator over feature existence;
extensions include combining multiple persistence and emergence
hypotheses~\cite{saavedra2025perpetua}. Across these methods, the model is either fit
from accumulated in-context observations or specified by hand from semantic class or
design heuristics.



\subsection{Foundation Models as Priors}
Large language and vision-language models exhibit temporal reasoning~\cite{chu2024timebench,yuan2025eocbench},
visually grounded multimodal reasoning~\cite{liu2024mmbench}, and embodied spatial
reasoning~\cite{graesser2026geminiroboticser16} capabilities that support commonsense reasoning about object use,
human activity, and scene context. A growing body of work turns these
capabilities into priors for downstream tasks, such as priors over likely
object co-occurrences for semantic search~\cite{zhou2023esc}
or over world states for planning~\cite{zhao2023llm}.
Closest to our setting, two recent methods derive persistence-related priors from
large language models (LLMs). Bogenberger et al.~\cite{bogenberger2026glasses} query an LLM for a
binary static/dynamic label per semantic class, which sets an object's stationarity score decay rate. Saavedra-Ruiz et al.~\cite{saavedra2026predictive} extend Perpetua~\cite{saavedra2025perpetua} with a prior derived from LLM predictions of an
object's location at a future time. In both, the language model contributes only a coarse prior rather than an instance- or scene-specific
one.



\subsection{Learned Visual Representations}
Large-scale foundation models such as DINOv2~\cite{oquab2023dinov2} provide
general-purpose visual features, which downstream methods commonly adapt by
fine-tuning or attaching a task-specific prediction head. More sophisticated
strategies train lightweight adapters on top of the frozen backbone~\cite{lu2024towards,adkins2025clover}.

A common way to obtain an object-level representation is masked average pooling, which
aggregates image features within an object's mask. Other
object-centric architectures form a query that cross-attends to image features.
DETR~\cite{carion2020detr} applies this to learned queries without
instance-specific inputs to generate detections;
SAM~3~\cite{carion2025sam3segmentconcepts} is prompted by points, boxes, masks, or
concepts to generate segmentations; and REN~\cite{khosla2025ren}, the closest to our
model, uses a point-prompted query over a frozen foundation backbone. PreSIST-Vis
builds on these ideas, pairing a frozen, lightly adapted DINOv2 backbone for open-world
generalization with a mask-pooled query that cross-attends to the patch tokens to produce
a per-object output.

\section{Preliminaries}
\label{sec:persistence_filter_prelim}

We build on the persistence filter of Rosen et al.~\cite{rosen2016towards}, adapting its survival model to object persistence (how long an object remains in its last observed pose). We briefly review relevant details for this work and refer the reader to~\cite{rosen2016towards} for the full recursive Bayesian estimator.




\subsection{Object Persistence as Survival}

Let $T \in [0,\infty)$ denote the survival time of an observed object instance after it is initialized, and define the persistence state at elapsed time $t$ as $X_t=\mathds{1}[t\leq T]$. Here, $X_t=1$ indicates that the object persists at time $t$ and $X_t=0$ indicates that it no longer occupies its last observed pose.

The robot does not observe $X_t$, instead obtaining a noisy detection $Y_t \in \{0,1\}$. A persistence-filter
model specifies
\begin{equation}
T \sim p_T(\cdot),
\qquad
Y_t \mid X_t \sim p_{Y_t}(\cdot \mid X_t),
\end{equation}
where $p_T$ is a prior density over survival times and $p_{Y_t}$ is the detector observation model, parameterized by a missed-detection probability $P_M$ and a false-alarm probability $P_F$.



\subsection{Survival Priors and Persistence Posterior}

The central modeling component of a persistence filter is the survival-time prior $p_T$, which encodes how long an observed object is expected to remain in place before incorporating observations. We represent this prior primarily through its survival function, $S_T(t)=p(T > t)$, the prior probability that the object persists at elapsed time $t$.



For query times $t \geq t_N$, given observations $\mathcal{Y}_{1:N}=\{(t_i,y_i)\}_{i=1}^N$, the persistence filter maintains the posterior
\begin{equation}
\begin{aligned}
p(X_t=1 \mid \mathcal{Y}_{1:N})
&=
p(T \geq t \mid \mathcal{Y}_{1:N})\\
&=
C(\mathcal{Y}_{1:N}) S_T(t),
\end{aligned}
\end{equation}
which is the probability that the object persists at elapsed time $t$ given the observations, with $C(\mathcal{Y}_{1:N})$ determined by the detector model and the observations up to time $t_N$. The filter thus separates two sources of information: an observation model for incorporating detections and a survival prior that governs how the belief decays without new data.


\subsection{Weibull Survival Priors}

We use Weibull distributions as parametric survival priors. These are widely used in survival analysis and have been used for object-persistence modeling in long-term robot navigation~\cite{symmank_spatio-temporal_2023}. A Weibull survival prior with shape parameter $k>0$ and scale parameter $\lambda>0$ has survival function
\begin{equation}
S_T(t;k,\lambda)
=
\exp\!\left[-\left(t/\lambda\right)^k\right],
\quad t \geq 0.
\end{equation}
The shape parameter $k$ controls the survival curve (constant, increasing, or decreasing decay rate for $k=1$, $k>1$, or $k<1$, respectively), and the scale parameter $\lambda$ sets the characteristic timescale of persistence. This two-parameter survival prior is flexible enough to model a range of persistence behaviors, yet still yields probabilities in closed form.

The Weibull distribution also admits a closed-form inverse survival function. For a survival probability level $s \in (0,1)$,
\begin{equation}
S_T^{-1}(s;k,\lambda)
=
\lambda\left[-\ln s\right]^{1/k}.
\end{equation}
Thus, $S_T^{-1}(s;k,\lambda)$ is the elapsed time at which the prior probability of persistence equals $s$, giving a survival-time quantile of the prior.

In this paper, PreSIST leverages the existing Bayesian filtering mechanism and focuses on estimating object-instance survival priors from object appearance and scene context.

\section{PreSIST}

\ifdefined\ifpresistverboseloss\else
  \expandafter\newif\csname ifpresistverboseloss\endcsname
\fi

\label{sec:presist}

PreSIST estimates an object-instance survival prior from a single observation.
As illustrated in Fig.~\ref{fig:presist_overview}, both variants share a common
interface that consumes an image and an instance segmentation mask and produces a
parametric survival prior. This prior initializes the persistence filter of
Sec.~\ref{sec:persistence_filter_prelim}, which maintains a belief over whether the
instance persists and updates it as new observations arrive. In the remainder of
this section, we first present the shared problem formulation, then describe the two
variants: the language-based PreSIST-Lang and the vision-only PreSIST-Vis.

\begin{figure*}[tbp]
    \vspace{0.5em}
    \centering
    \includegraphics[width=0.65\textwidth]{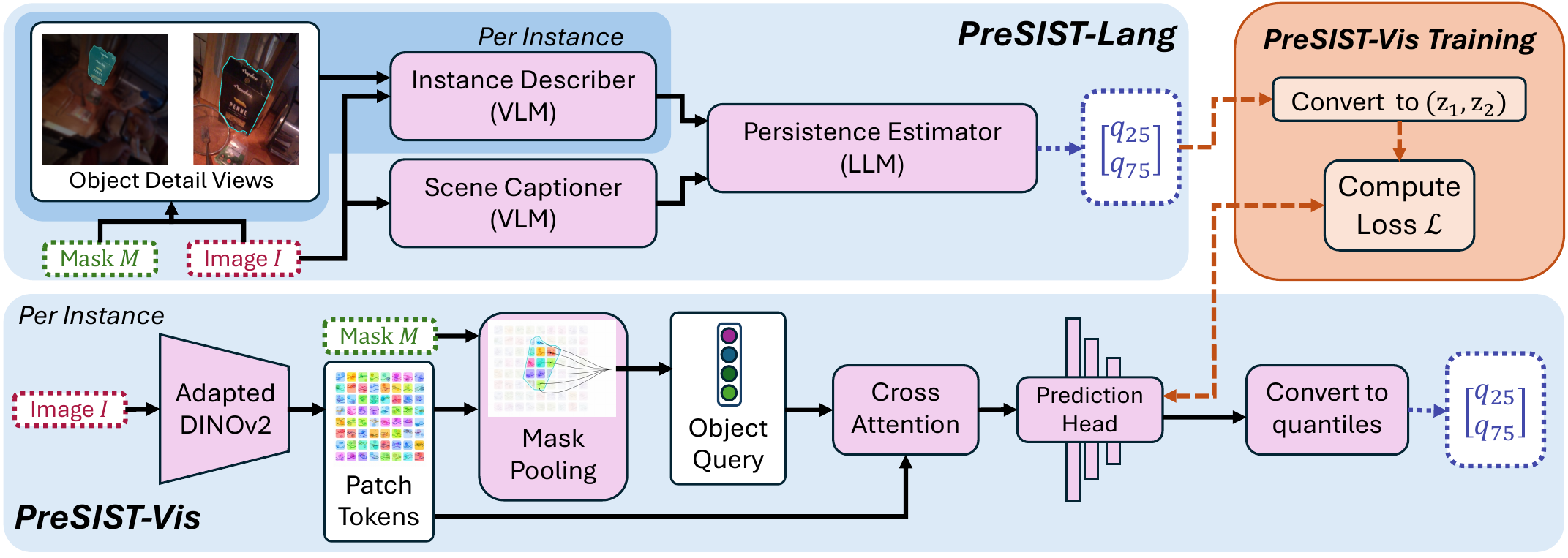}
    \caption{The two PreSIST predictor instantiations. \emph{PreSIST-Lang} (top)
    issues two vision-language model queries to caption image $I$ and describe each instance from $I$ plus two object detail
    views of mask $M$; given these outputs, a language model produces
    survival-time quantiles $(q_{25}, q_{75})$.
    \emph{PreSIST-Vis} (bottom) encodes $I$ into patch tokens, mask-pools them under $M$ into an object query that
    cross-attends to the patch tokens and regresses the quantiles. In training, PreSIST-Lang
    quantiles become log-space targets $(z_1, z_2)$ for loss $\mathcal{L}$. Dotted inputs and outputs mark connections to the overall pipeline (Fig.~\ref{fig:presist_overview}).}
    \label{fig:presist_methods}
    \vspace{-1.5em}
\end{figure*}

\subsection{Problem Formulation}
\label{subsec:presist_overview}

Each PreSIST variant defines a predictor $f$ that maps an image $I$ and an
instance mask $M$ to two quantiles of the instance's survival time $T$,
\begin{equation}
    f(I, M) = (q_{25}, q_{75}),
\end{equation}
where the quantile $q_{\tau}$ is the elapsed time by which a fraction $\tau$ of
comparable instances are expected to have left their last observed pose.
Equivalently, $p(T \leq q_{\tau}) = \tau$, or $S_T(q_{\tau}) = 1-\tau$. We represent the instance's survival-time prior $p_T$ with this pair of quantiles
rather than its mean or its native shape and scale parameters $(k, \lambda)$, as quantiles are directly interpretable, both
for the language model that produces them and for downstream analysis, while a pair
still uniquely determines $(k, \lambda)$. We use the $25$th and
$75$th percentiles because they bound the central $50\%$ of probability mass
while avoiding the distribution's hard-to-estimate tails.

We map the predicted quantiles to a Weibull survival prior in closed form. Since
the inverse survival function gives $q_{\tau} = \lambda\left[-\ln(1-\tau)\right]^{1/k}$,
the pair $(q_{25}, q_{75})$ determines $(k,\lambda)$:
\begin{equation}
    k = \frac{a_{0.75} - a_{0.25}}{\ln q_{75} - \ln q_{25}},
    \qquad
    \ln \lambda = \ln q_{25} - \frac{a_{0.25}}{k},
    \label{eq:quantile_to_weibull}
\end{equation}
where $a_\tau = \ln\!\left(-\ln(1-\tau)\right)$. These parameters initialize the persistence filter with the
object-instance survival prior $S_T(t;k,\lambda)$, leaving its belief update unchanged. PreSIST thus contributes
a scene-conditioned survival prior while preserving the generality and
efficiency of the established filtering framework. Because both variants share the quantile
representation and its mapping to a Weibull prior, they differ only in how they predict $(q_{25}, q_{75})$ from $(I, M)$.

\subsection{PreSIST-Lang}
\label{subsec:presist_lang}

PreSIST-Lang predicts the survival-time quantiles $(q_{25}, q_{75})$ for an
instance by prompting a VLM, exploiting the
commonsense knowledge such models carry about object use, human activity, and how
scenes evolve over time. As depicted in Fig.~\ref{fig:presist_methods}, rather than query a single model for persistence directly
from the image, we decompose the prediction into two independent description
steps---one scene-level and one instance-level, which can be computed in
parallel---followed by a text-only reasoning step that combines them to produce the
quantiles. This decomposition separates concerns across the steps so that each
focuses on a distinct source of evidence: the surrounding scene context and
activity at the scene level, and the specific instance's appearance and state at
the instance level. The final reasoning step
draws on both, letting it weigh how scene-level processes apply to this particular
instance rather than relying on scene context or object category alone.

\subsubsection{Scene Description}
A VLM describes the full image $I$, producing a scene-level
caption that emphasizes the environmental processes and activity relevant to
persistence, such as traffic flow, dining, or retail turnover, rather than
detailed appearance. This caption is computed once per image and shared across
all associated instances.

\subsubsection{Instance Description}
For each instance, a VLM produces a natural-language description
of the object and its interactions with its surroundings from $I$ and mask $M$. To convey
the object of interest while preserving context, we present the model with the full image and two object-specific views, depicted in Fig.~\ref{fig:presist_methods}, that highlight the instance: a crop around the object and a full image with the background blurred and dimmed, both with a mask outline. The description identifies the object and reports the visual
evidence relevant to its possible motion or persistence,
including its support and contact relationships, nearby entities, and any signs of
ongoing or likely future movement.


\subsubsection{Persistence Reasoning}
A text-only model receives the scene description and the instance
description, then produces the survival-time quantiles $(q_{25}, q_{75})$ for each mask queried for the image. Because
this step operates purely on text, its temporal reasoning is grounded in the
structured visual evidence extracted by the earlier steps. The prompt specifies the desired reasoning process, guiding the model to identify processes that could change the instance's state
and their \cutAAA{characteristic }timescales; weigh the factors that influence the likelihood of these processes for this instance; and then produce
$(q_{25}, q_{75})$. The full prompts are available in
our codebase.

\subsection{PreSIST-Vis}
\label{subsec:presist_vis}

PreSIST-Vis is a vision-only predictor that estimates the survival-time quantiles
$(q_{25}, q_{75})$ for an instance directly from the image $I$ and mask $M$ in a
single forward pass. It is trained to reproduce the outputs of PreSIST-Lang, retaining context-sensitive estimates without
online large-model inference at deployment time.

\subsubsection{Training Data}
We query PreSIST-Lang to generate persistence pseudo-labels for training
PreSIST-Vis, a process that requires images and instance masks. We draw
images from a diverse set of indoor and outdoor
datasets---COCO~\cite{lin2015microsoft},
OpenLORIS-Scene~\cite{shi2019openlorisscene},
OpenLORIS-Location~\cite{li2020RaPNet}, KITTI~\cite{Geiger2012CVPR},
EPIC-Kitchens~\cite{Damen2022RESCALING}, UT
CODa~\cite{zhang2023towards}, SKU110k~\cite{goldman2019dense},
SUN397~\cite{Xiao:2010}, and the UFPR subset of PKLot~\cite{de2015pklot}---so that
the pseudo-labels span a wide range of objects, scenes, and contexts. To obtain masks, we first query a vision-language model
(Qwen3-VL-8B-Thinking~\cite{bai2025qwen3vl})
for the object classes present in each image, then prompt
SAM~3~\cite{carion2025sam3segmentconcepts} with each class to segment corresponding
instances. Running PreSIST-Lang on the resulting image--mask pairs yields a
$(q_{25}, q_{75})$ pseudo-label for every instance, which serves as the
target for PreSIST-Vis.

\subsubsection{Architecture}
We implement PreSIST-Vis as a novel mask-conditioned cross-attention network, depicted in Fig.~\ref{fig:presist_methods}. A
frozen DINOv2 ViT-L/14 backbone~\cite{oquab2023dinov2}, augmented with lightweight trainable
adapters~\cite{lu2024towards, adkins2025clover}, encodes the image $I$ into a grid of
patch-token features. The patch features within the mask $M$ are aggregated into
a single object query vector. The query then cross-attends to all patch tokens, gathering the scene
context most relevant to that instance, and a prediction head maps the attended feature to the model outputs. Conditioning the
attention on the instance query while attending over the full image lets the
network combine object-specific appearance with surrounding context in a single
pass, mirroring the two evidence sources that PreSIST-Lang draws on across its
description steps.

\subsubsection{Training Objective}
Rather than regress $(q_{25}, q_{75})$ directly, which is difficult as
survival times span many orders of magnitude and must satisfy $q_{25} > 0$ and
$q_{75} \ge q_{25}$, PreSIST-Vis predicts estimates $\hat{z}_1, \hat{z}_2$ of two
transformed targets, $z_1 = \log q_{25}$ and $z_2 = \log(q_{75}/q_{25} - 1)$.\footnote{The network also has a
third output head trained to reproduce PreSIST-Lang-generated confidence estimates with a binary
cross-entropy term; this estimate is not otherwise used for
persistence prediction in this work.} This parameterization is scale-invariant,
and decoding $q_{25} = e^{\hat{z}_1}$ and $q_{75} = q_{25}\,(1 + e^{\hat{z}_2})$
recovers a valid ordered pair by construction. We standardize $z_1$ and $z_2$ to
zero mean and unit variance using statistics computed over the training set, and apply a Huber loss $H_\delta$, with per-target threshold $\delta$:
\begin{equation}
    \mathcal{L} = H_{\delta_1}\!\bigl(\hat{z}_1, z_1\bigr)
                + H_{\delta_2}\!\bigl(\hat{z}_2, z_2\bigr).
    \label{eq:presist_vis_loss}
\end{equation}
We also sample instances so that each image
contributes equally, preventing images with many instances from dominating the
loss. At inference, predictions are converted to $(q_{25}, q_{75})$,
yielding the same interface as PreSIST-Lang.

\section{Experimental Results}
\label{sec:result}

\begin{figure*}[!t]
    \vspace{0.5em}
    \centering
    \includegraphics[width=0.9\textwidth]{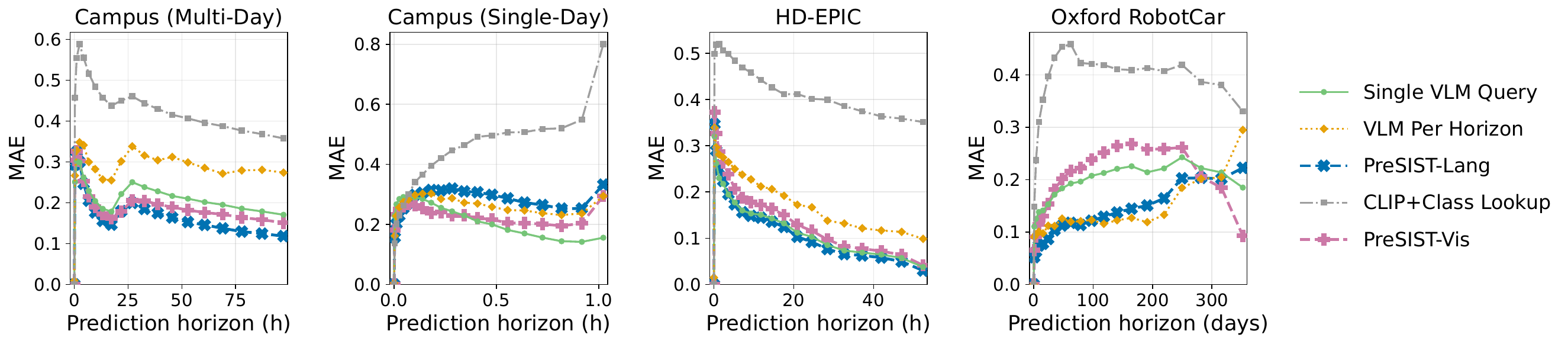}
    \caption{Zero-shot persistence prediction MAE over varying prediction horizons across all evaluation domains; lower is better.}
    \label{fig:zero-shot-pred-horizon}
    \vspace{-1.5em}
\end{figure*}

We aim to answer the following questions in our evaluation:
\begin{enumerate*}[label=(\roman*)]
    \item How accurately can our method generate persistence estimates without task-specific training data?
    \item How does our method using open-world semantic and contextual information compare to methods that train on in-context data?
    \item What is the impact of each component of our method on its persistence prediction performance?
    \item Can open-world persistence estimates improve a downstream task, specifically long-term visual relocalization?
\end{enumerate*}
We describe the common experimental setup then address each
question in turn.

\subsection{Implementation Details}

Unless otherwise noted, we use the Weibull survival prior. PreSIST-Lang and other VLM-based methods query a locally hosted Gemma~4 31B instruction-tuned model~\cite{google_gemma_4_31b_it_2026}. PreSIST-Vis is trained with AdamW at a learning rate of $1\times10^{-5}$, batch size $32$, and $64$k sampled instances per epoch. We train for up to $100$ epochs, stopping if validation performance does not improve for $10$ epochs. PreSIST-Vis and its ablations are each trained and evaluated on a single NVIDIA H100.

At inference, each method receives an image and instance segmentation mask for a single object and outputs persistence parameters that seed a new persistence filter. Across all evaluations, we draw $1000$ query masks per domain, sampled throughout each object's track and balanced across tracks so that long- and short-lived objects are represented roughly equally. We query each filter at $20$ prediction horizons (future times measured relative to the observation), scaled per domain to its distribution of object lifetimes. Each track's first and last observed times, together with the first observed absence when available, bound the true disappearance time and provide the present/absent label at each horizon; horizons that fall in the unknown window between the last observation and the first observed absence are excluded from the metrics.

\subsection{Zero-Shot Persistence Estimation}

\textbf{Data and Annotations~~}Evaluating open-world persistence prediction requires visual data with instance segmentation for varied contexts and object classes, capturing object appearance and disappearance in the wild. As no existing dataset provides this, we contribute a dataset of stationary object track annotations that capture, for each object, a set of observations (timestamped instance segmentation masks) along with the times it was last seen present and first seen absent. We use three video sources: footage we collected on the \blindtext{UT Austin} campus, and subsets of the Oxford RobotCar~\cite{RobotCarDatasetIJRR} and HD-EPIC~\cite{perrett2025hdepic} datasets, which together yield four evaluation domains spanning wide-ranging timescales. Oxford RobotCar is collected over a year and HD-EPIC over several consecutive days in each kitchen. We divide our campus footage into two groups: \emph{Campus (Single-Day)} comprises two hour-long sessions at a building lobby and at outdoor campus tables, while \emph{Campus (Multi-Day)} comprises a kitchenette recorded for at least $100$ minutes per day for $5$ days and a street scene captured in two 30+ minute sessions two weeks apart. We segmented objects with SAM~3~\cite{carion2025sam3segmentconcepts} and manually annotated presence and absence, yielding more than $250$ object tracks and over $6500$ segmentation masks.

\textbf{Experimental Setup~~}To assess the performance of our methods and others, we use balanced accuracy with a presence probability threshold of $0.5$ and mean absolute error (MAE) comparing the persistence probability with ground truth existence ($0$ for absent, $1$ for present), following~\cite{saavedra2025perpetua}. We also measure the average time per query in seconds.
No existing method produces informative open-world zero-shot priors, so we construct three baselines spanning the natural strategies: two that query a VLM and one vision-only; we report these categories in separate groups as they differ greatly in runtime. In \emph{Single VLM Query}, a single query is made to the VLM to obtain persistence parameters for each object detection of interest; in \emph{VLM Per Horizon}, the VLM is instead queried directly for the persistence probability at each prediction horizon. Unlike the filter-based methods, this baseline queries a VLM per horizon, so its reported time sums over horizons and would grow with additional queries. For the vision-only baseline, we precompute a mapping from semantic class to persistence parameters by querying an LLM once for every semantic class in COCO-Stuff~\cite{caesar2018cvpr}; at inference, we use CLIP~\cite{radford2021clip} to assign each detection its closest class and look up the corresponding parameters (\emph{CLIP+Class Lookup}). To isolate the quality of the survival priors, each prediction uses only the single query image, without additional positive or negative observations.

\begin{table*}[!t]
\vspace{0.5em}
\centering
\scriptsize
\caption{Persistence prediction MAE, balanced accuracy (B-Acc), and average per-query time for zero-shot estimators. In each column, the best value within each category is bolded.}
\label{tab:zero_shot_acc}
\begin{tabular}{llrrr|rrr|rrr|rrr}
    & & \multicolumn{3}{c|}{Campus (Multi-Day)}
      & \multicolumn{3}{c|}{Campus (Single-Day)}
      & \multicolumn{3}{c|}{HD-EPIC}
      & \multicolumn{3}{c}{Oxford RobotCar} \\ \cline{3-14}
    & & \multicolumn{1}{c}{MAE}
      & \multicolumn{1}{c}{B-Acc}
      & \multicolumn{1}{c|}{Time (s)}
      & \multicolumn{1}{c}{MAE}
      & \multicolumn{1}{c}{B-Acc}
      & \multicolumn{1}{c|}{Time (s)}
      & \multicolumn{1}{c}{MAE}
      & \multicolumn{1}{c}{B-Acc}
      & \multicolumn{1}{c|}{Time (s)}
      & \multicolumn{1}{c}{MAE}
      & \multicolumn{1}{c}{B-Acc}
      & \multicolumn{1}{c}{Time (s)} \\ \hline
    \multirow{3}{*}{VLM}
    & Single VLM Query
    & 0.207 & 0.848 & \textbf{3.179} & \textbf{0.219} & \textbf{0.807} & \textbf{3.371} & 0.140 & 0.827 & \textbf{3.906} & 0.175 & 0.861 & \textbf{7.671} \\

    & VLM Per Horizon
    & 0.278 & 0.827 & 8.770 & 0.247 & 0.756 & 9.755 & 0.195 & \textbf{0.836} & 8.646 & 0.124 & 0.882 & 8.532 \\

    & PreSIST-Lang
    & \textbf{0.175} & \textbf{0.865} & 7.540 & 0.258 & 0.708 & 7.900 & \textbf{0.139} & 0.834 & 8.550 & \textbf{0.118} & \textbf{0.904} & 8.430 \\
    \hline
    \multirow{2}{*}{Vision-Only}
    & CLIP+Class Lookup
    & 0.418 & 0.691 & \textbf{0.016} & 0.376 & 0.513 & \textbf{0.016} & 0.414 & 0.711 & \textbf{0.016} & 0.359 & 0.656 & \textbf{0.015} \\

    & PreSIST-Vis
    & \textbf{0.193} & \textbf{0.847} & 0.042 & \textbf{0.218} & \textbf{0.768} & 0.042 & \textbf{0.165} & \textbf{0.810} & 0.040 & \textbf{0.181} & \textbf{0.852} & 0.041 \\
\end{tabular}
\vspace{-1.5em}
\end{table*}

\begin{figure}[tb]
    \centering
    \setlength{\tabcolsep}{2pt}
    \renewcommand{\arraystretch}{0.7}
    \begin{tabular}{cc}
        \includegraphics[width=0.3\columnwidth]{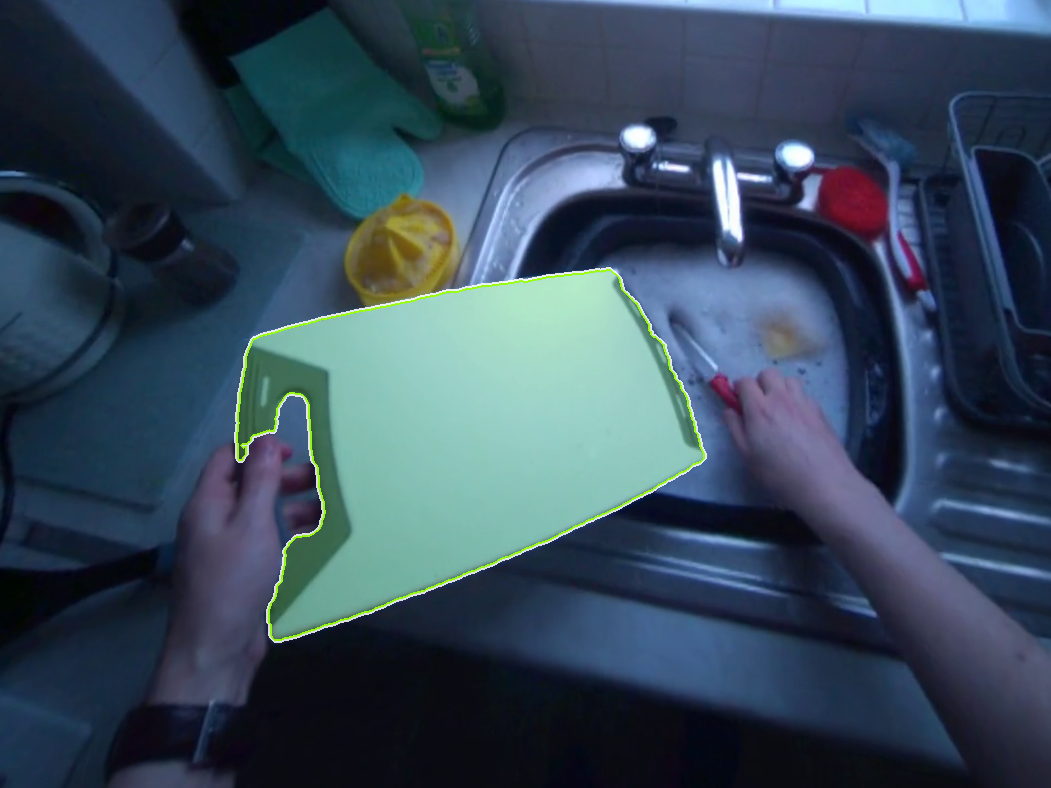} &
        \includegraphics[width=0.3\columnwidth]{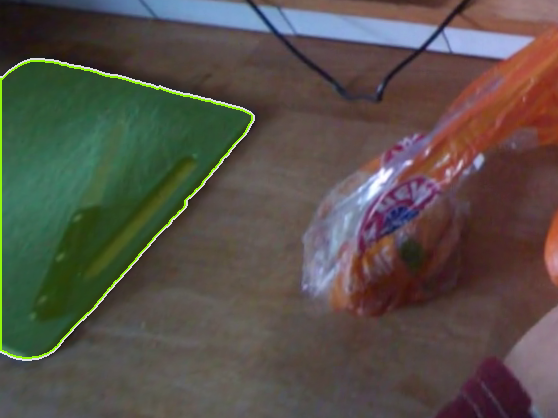} \\
        {\footnotesize (a) board, in hand} & {\footnotesize (b) board, on counter} \\[1.5pt]
        \includegraphics[width=0.3\columnwidth]{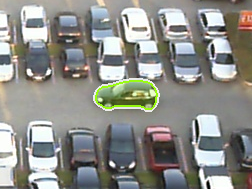} &
        \includegraphics[width=0.3\columnwidth]{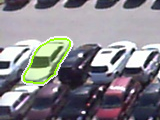} \\
        {\footnotesize (c) car, moving} & {\footnotesize (d) car, parked} \\
    \end{tabular}

    \vspace{3pt}
    \definecolor{durA}{HTML}{EFF3FF} 
    \definecolor{durB}{HTML}{C6DBEF} 
    \definecolor{durC}{HTML}{9ECAE1} 
    \definecolor{durD}{HTML}{6BAED6} 
    \providecommand{\dl}[2]{\cellcolor{dur#1}#2}
    {\scriptsize
    \setlength{\tabcolsep}{2.5pt}
    \renewcommand{\arraystretch}{1.1}
    \begin{tabular}{ll cc | cc}
        & & \multicolumn{2}{c|}{Cutting board} & \multicolumn{2}{c}{Car} \\ \cline{3-6}
        & Method & (a) & (b) & (c) & (d) \\ \hline
        \multirow{2}{*}{VLM}
        & Single VLM Query    & \dl{A}{0.0--0.7s}   & \dl{A}{0.0--0.7s}  & \dl{C}{1.0--8.0h} & \dl{C}{1.0--8.0h} \\
        & PreSIST-Lang        & \dl{A}{0.0--0.4s}   & \dl{B}{10min--2.0h} & \dl{A}{0.4--20s}  & \dl{C}{1.0--6.0h} \\ \hline
        \multirow{2}{*}{\shortstack[l]{Vision-\\Only}}
        & CLIP+Class Lookup   & \dl{C}{15min--2.0d} & \dl{C}{15min--2.0d} & \dl{C}{1.0--8.0h} & \dl{C}{1.0--8.0h} \\
        & PreSIST-Vis         & \dl{A}{0.1--1.5s}   & \dl{B}{14min--3.3h} & \dl{B}{1.8--16min} & \dl{C}{47min--5.7h} \\
    \end{tabular}}

    \caption{Persistence predictions for two object classes, each in
    a moving and stationary context: a cutting board being washed (a) vs.\ resting on the counter (b), and a car in a driving
    lane (c) vs.\ in a parking spot (d). The instance masks are shown in green. The table reports each method's predicted persistence interval
    $[q_{25}, q_{75}]$, with each cell
    shaded by its predicted duration (darker = longer).}
    \label{fig:qualitative}
\end{figure}

\begin{figure}[tb]
    \centering
    \includegraphics[width=0.9\columnwidth]{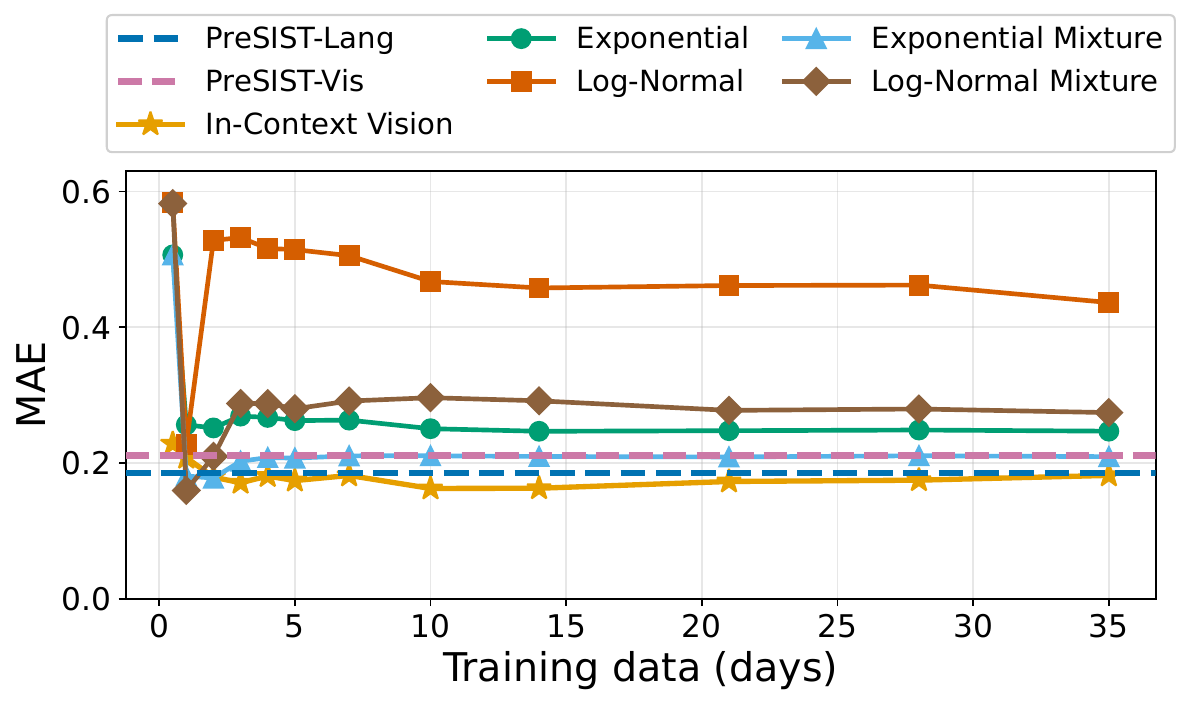}
    \caption{Persistence prediction MAE versus the amount of in-context training data on PKLot (PUCPR); lower is better.}
    \label{fig:training_set_size_variation}
\end{figure}

\textbf{Results~~}Fig.~\ref{fig:zero-shot-pred-horizon} shows the prediction MAE across varying prediction horizons for the Campus Multi-Day and Single-Day sequences, HD-EPIC, and Oxford RobotCar, and Table~\ref{tab:zero_shot_acc} reports the MAE, balanced accuracy, and prediction time per query for all approaches on all domains. Among the VLM-based methods, PreSIST-Lang attains the best or near-best balanced accuracy and MAE on the multi-day, HD-EPIC, and RobotCar domains, outperforming the per-horizon and single-query VLM baselines. PreSIST-Lang's per-query time is somewhat longer than the unstructured VLM baseline; this cost is acceptable because it is intended to run offline, where its accuracy advantage matters more than the speed difference. Among the vision-only methods, PreSIST-Vis dominates the CLIP+Class Lookup baseline by a wide margin, confirming that context and instance-specific reasoning are more informative than a class-only prior. Most notably, PreSIST-Vis incurs only a small drop in performance relative to PreSIST-Lang across domains while running in roughly $0.04$ seconds, making it the only approach suitable for effective online use. The single-day domain is the hardest case and the one place PreSIST-Lang lags, likely due to subtle differences in short-horizon prediction. PreSIST-Vis nonetheless remains competitive, tying the Single VLM Query baseline for the best MAE. Finally, Fig.~\ref{fig:qualitative} shows two semantic classes, a cutting board and a car, each observed in a short-lived and a long-lived context (the board held at the sink versus resting on the counter, and the car moving versus parked), together with each method's predicted persistence interval $[q_{25}, q_{75}]$. The class-prior CLIP+Class Lookup baseline and the Single VLM Query baseline return nearly the same estimate in both contexts, whereas PreSIST-Lang and PreSIST-Vis adapt their estimates to the instance and its context. PreSIST-Vis still somewhat overestimates the moving car, predicting a duration of minutes rather than the seconds-scale estimate of PreSIST-Lang, but it nonetheless separates the moving and parked cases by a wide margin.

\subsection{PreSIST vs In-Context Training Methods}

\begin{table*}[!t]
\vspace{0.5em}
\centering
\scriptsize
\caption{Persistence prediction MAE, balanced accuracy (B-Acc), and per-query time for our primary methods and their ablations. Within each category (VLM, Vision-Only), the best value per column is bolded and the second best is underlined.}
\label{tab:ablations}
\setlength{\tabcolsep}{5.5pt}  
\begin{tabular}{llrrr|rrr|rrr|rrr}
    & & \multicolumn{3}{c|}{Campus (Multi-Day)}
      & \multicolumn{3}{c|}{Campus (Single-Day)}
      & \multicolumn{3}{c|}{HD-EPIC}
      & \multicolumn{3}{c}{Oxford RobotCar} \\ \cline{3-14}
    & & \multicolumn{1}{c}{MAE}
      & \multicolumn{1}{c}{B-Acc}
      & \multicolumn{1}{c|}{Time (s)}
      & \multicolumn{1}{c}{MAE}
      & \multicolumn{1}{c}{B-Acc}
      & \multicolumn{1}{c|}{Time (s)}
      & \multicolumn{1}{c}{MAE}
      & \multicolumn{1}{c}{B-Acc}
      & \multicolumn{1}{c|}{Time (s)}
      & \multicolumn{1}{c}{MAE}
      & \multicolumn{1}{c}{B-Acc}
      & \multicolumn{1}{c}{Time (s)} \\ \hline
    \multirow{3}{*}{VLM}
    & PreSIST-Lang
    & 0.175 & \underline{0.865} & 7.540 & \underline{0.258} & \underline{0.708} & 7.900 & 0.139 & \underline{0.834} & 8.550 & 0.118 & \underline{0.904} & 8.430 \\

    & Weibull
    & \underline{0.169} & 0.859 & \underline{7.420} & 0.266 & 0.676 & \underline{7.774} & \textbf{0.126} & \textbf{0.835} & \underline{8.404} & \textbf{0.094} & 0.903 & \underline{8.280} \\

    & Gemini Robotics-ER 1.6
    & \textbf{0.124} & \textbf{0.900} & \textbf{5.747} & \textbf{0.234} & \textbf{0.739} & \textbf{6.275} & \underline{0.131} & 0.809 & \textbf{7.340} & \underline{0.107} & \textbf{0.913} & \textbf{7.226} \\
    \hline
    \multirow{5}{*}{Vision-Only}
    & PreSIST-Vis
    & \textbf{0.193} & \underline{0.847} & 0.042 & \underline{0.218} & \underline{0.768} & 0.042 & 0.165 & 0.810 & 0.040 & \underline{0.181} & \textbf{0.852} & 0.041 \\

    & Masked Region Only
    & 0.236 & 0.837 & 0.043 & 0.240 & 0.733 & 0.044 & \underline{0.154} & \underline{0.816} & 0.041 & \textbf{0.181} & \underline{0.848} & \underline{0.040} \\

    & Frozen Backbone
    & 0.220 & 0.838 & 0.040 & 0.227 & 0.757 & \underline{0.040} & 0.183 & 0.778 & \textbf{0.036} & 0.189 & 0.839 & \textbf{0.037} \\

    & No Pretrained Backbone
    & 0.385 & 0.719 & \underline{0.040} & 0.377 & 0.546 & 0.043 & 0.275 & 0.689 & \underline{0.040} & 0.367 & 0.685 & 0.040 \\

    & Dense Prediction
    & \underline{0.201} & \textbf{0.855} & \textbf{0.031} & \textbf{0.214} & \textbf{0.774} & \textbf{0.033} & \textbf{0.153} & \textbf{0.829} & 0.042 & 0.224 & 0.817 & 0.040 \\
\end{tabular}
\vspace{-1.5em}
\end{table*}

\textbf{Experimental Setup~~}We compare PreSIST-Lang and PreSIST-Vis against methods that train on in-context data, using the PUCPR subset of PKLot~\cite{de2015pklot}, which captures cars in a single parking lot over many days. Our methods need no PKLot-specific training; we ask how much in-context training data the baselines need to match this zero-shot performance. We augment PKLot labels with cars outside the annotated spaces, including transient ones with different persistence patterns, by running a VisDrone-fine-tuned YOLOv8~\cite{jocher2023yolov8,zhu2021visdrone} on the parking lot images. We identify individual vehicle lifetimes by splitting each location's occupancy labels at gaps in vehicle presence, and manually verify the tracks, rejecting spurious detections and splitting tracks in which one vehicle departed and another arrived between captures. Per-observation instance masks are generated with SAM~3~\cite{carion2025sam3segmentconcepts}. We divide the tracks by date into $50\%/20\%/30\%$ train/validation/test splits. We compare against the exponential and log-normal distributions of Perpetua~\cite{saavedra2025perpetua}, each in both single-component and mixture form, fitting separate parameters for each location. We omit the location-based emergence model in~\cite{saavedra2025perpetua}, as our task is object-focused. We also compare against a vision-only model that uses PreSIST-Vis's architecture but regresses a single mean lifetime to match observed persistence, with an exponential prior (\emph{In-Context Vision}). We train each in-context baseline on varying amounts of the training split, from $0.5$ to $35$ days, and evaluate on the held-out test split. As the mixture models benefit from observations to distinguish between persistence parameter sets, we provide all methods with up to $30$ minutes of observations after the initial observation, enforcing a gap between the last observation and prediction horizon.

\textbf{Results~~}
Fig.~\ref{fig:training_set_size_variation} plots prediction MAE against the amount of in-context training data; our training-free methods appear as horizontal lines. With little training data, the data-driven baselines underperform both PreSIST variants. The baselines improve with more data, narrowing the gap. However, even in this limited-diversity setting, only the exponential mixture and In-Context Vision models outperform PreSIST-Vis, and PreSIST-Lang consistently outperforms all but In-Context Vision. This highlights the value of open-world priors when in-context data is scarce, and the continued competitiveness of PreSIST even with in-context training.

\subsection{Persistence Model Ablations}

We group our ablations to match the two method families in Table~\ref{tab:ablations}. For the VLM-based PreSIST-Lang, we vary the survival-prior parameterization by directly predicting Weibull parameters instead of deriving them from quantiles. We also evaluate using Gemini Robotics-ER 1.6~\cite{graesser2026geminiroboticser16} as a stronger but proprietary alternative to the open-weight Gemma VLM. For the vision-only PreSIST-Vis, we ablate the architecture and training: using only the masked object region (\emph{Masked Region Only}), removing the adapters so the pretrained backbone is frozen (\emph{Frozen Backbone}), training the backbone from scratch (\emph{No Pretrained Backbone}), and predicting per-pixel and then pooling, in place of the mask-pooled cross-attention query (\emph{Dense Prediction}).

All VLM variants are competitive, but the proprietary Gemini Robotics-ER backbone is strongest overall: it is the fastest and has lower MAE than PreSIST-Lang on every domain, with the best balanced accuracy on three of the four. Directly predicting the Weibull parameters instead of quantiles attains the lowest MAE on HD-EPIC and RobotCar, but is slightly worse on balanced accuracy. Within the vision-only family, forgoing the pretrained backbone dramatically decreases performance, confirming that large-scale visual pretraining is essential for generalization. Using a fully frozen backbone also yields consistently worse performance, though less pronounced than not pretraining. The remaining vision-only ablations have less impact; PreSIST-Vis nonetheless performs best overall across all domains, while each other variant degrades on at least one.

\subsection{Downstream Task: Long-Term Relocalization}

\textbf{Experimental Setup~~}
Finally, we test whether open-world persistence estimates improve a downstream task: long-term visual relocalization against a previously built map. We evaluate on the ExMaps~\cite{rotsidis2021exmaps} grocery store dataset, where regular product turnover changes the scene over time, challenging relocalization. We use COLMAP~\cite{schoenberger2016sfm} to reconstruct a map from one sequence. We then localize query images against this map, drawn from sequences captured twice daily over the $5$ days following mapping; their ground-truth poses come from COLMAP's registration of the query frames. Each 3D map point is associated with a persistence filter whose Weibull prior is derived from the median PreSIST estimates of the segmented regions containing the point. Before pose estimation, we discard every map point whose survival probability at the query image time falls below $0.25$, then run EPnP-based RANSAC on the surviving correspondences. We compare against RANSAC with no filtering.

{\setlength{\tabcolsep}{3pt}
\begin{table}[!tb]
\centering
\scriptsize
\caption{Total survival prior construction time over $81{,}595$ points from $339$ images, RANSAC efficiency metrics, and localization accuracy for long-term relocalization on the ExMaps dataset, where query images are localized against a map built on an earlier day, with and without filtering map points by PreSIST predictions.}
\label{tab:retail_ransac_persistence}
\begin{tabular}{lrrrrrr}
\hline
Filtering & $S_T$ time\,$\downarrow$ & Query time (s)\,$\downarrow$ & Inl. \%\,$\uparrow$ & Iter.\,$\downarrow$ & $t$ (m)\,$\downarrow$ & $R$ ($^\circ$)\,$\downarrow$ \\
\hline
None                & --       & 0.224 & 31.96 & 1221 & 0.131 & 6.67 \\
PreSIST-Lang    & 2.3 h    & 0.101 & 42.89 &  731 & 0.115 & 4.86 \\
PreSIST-Vis   & 21 min & 0.113 & 39.78 &  799 & 0.112 & 5.17
\end{tabular}
\end{table}}

\begin{figure}[tb]
    \centering
    \includegraphics[width=0.9389\columnwidth]{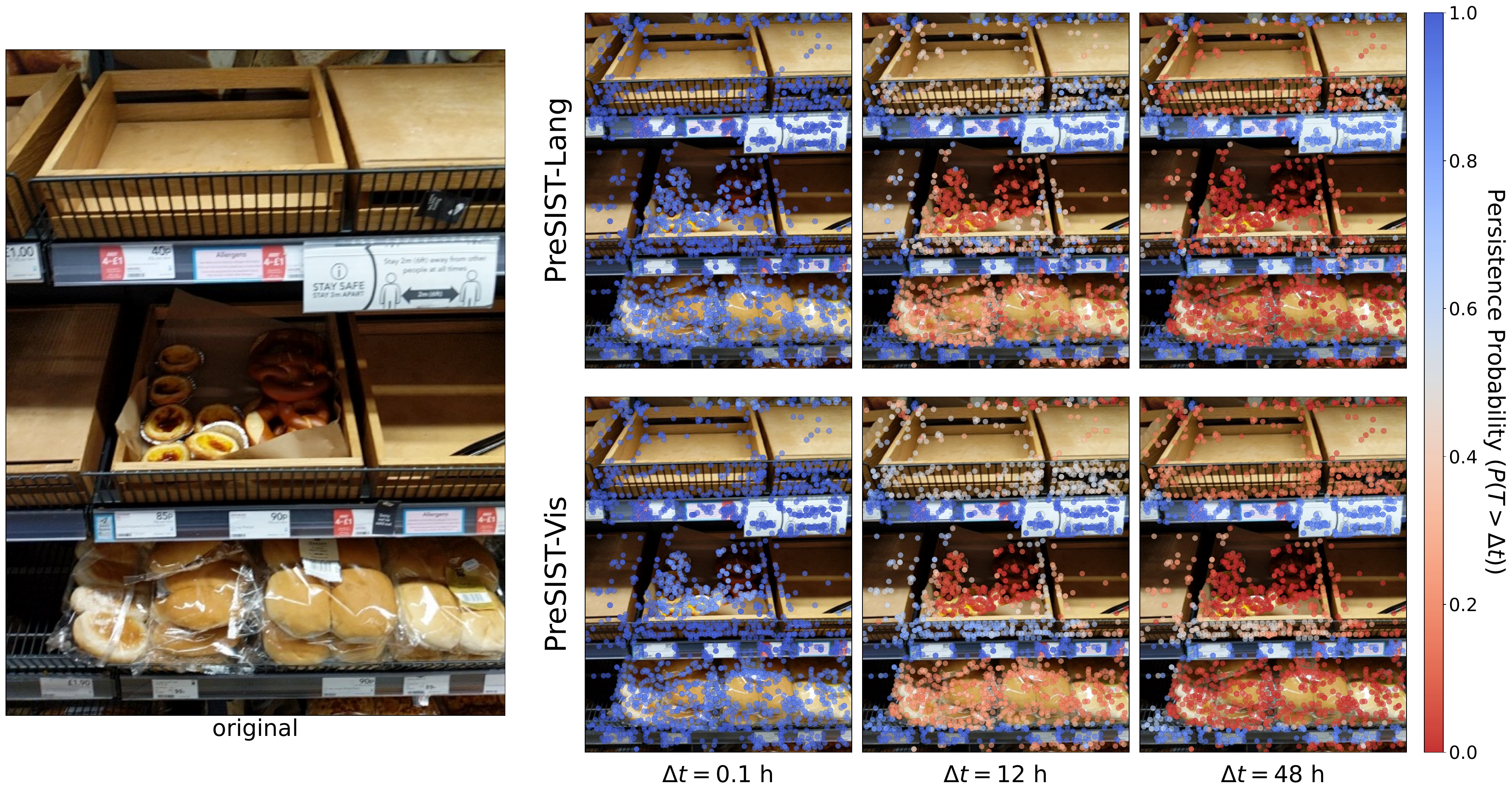}
    \caption{A mapped image colored per-pixel by the estimated probability that the map feature still exists at the specified query time.}
    \label{fig:reloc_heatmap}
\end{figure}

\textbf{Results~~}
Table~\ref{tab:retail_ransac_persistence} reports relocalization quality and computational efficiency. We report PreSIST's one-time cost to infer survival priors for all map points, and the average per-query pose estimation time. For both PreSIST-Lang and PreSIST-Vis, filtering out low-persistence map points raises the inlier ratio (Inl.\ \%), reduces RANSAC iterations (Iter.), and lowers translation ($t$ (m)) and rotation error ($R$ ($^\circ$)) relative to the unfiltered baseline. Fig.~\ref{fig:reloc_heatmap} visualizes the persistence probability of map points across prediction horizons: probabilities are similar at $0.1$ hours, but as the horizon grows, points on shelving stay highly persistent while those on higher-turnover products drop sharply.


\section{Conclusion}
\label{sec:conclusion}

In this paper, we propose PreSIST, a method for zero-shot open-world object persistence prediction that estimates survival priors from object properties and context. PreSIST combines commonsense temporal and behavioral reasoning from large VLMs with persistence filters, enabling probabilistic prediction at arbitrary future time horizons. We introduce two variants: PreSIST-Lang, which uses VLMs to estimate persistence quantiles, and PreSIST-Vis, a vision-only model trained from PreSIST-Lang pseudo-labels for faster deployment. Our results show that PreSIST improves zero-shot persistence prediction, with PreSIST-Lang achieving the strongest accuracy and PreSIST-Vis matching it closely at a fraction of the runtime.

Future work could extend PreSIST in several directions. Short videos rather than single images could improve persistence estimates with stronger cues about ongoing activity and imminent motion. Richer survival models, such as piecewise or mixture functions, could better capture complex object dynamics, \eg{} kitchen objects that move more frequently around meal times. Finally, PreSIST-Vis could be specialized to a deployment environment by generating additional PreSIST-Lang pseudo-labels on target environment images or fine-tuning on a small amount of in-domain persistence data, combining open-world priors with environment-specific experience for improved long-term estimation.




\bibliographystyle{IEEEtran}
\bibliography{references}  

\clearpage

\end{document}